\documentclass[conference]{IEEEtran}
\IEEEoverridecommandlockouts
\usepackage{cite}
\usepackage{amsmath,amssymb,amsfonts}
\usepackage{algorithmic}
\usepackage{graphicx}
\usepackage{multirow}
\usepackage{multicol}
\usepackage{graphicx}
\usepackage{adjustbox}
\usepackage{mathtools}
\usepackage{threeparttable}
\usepackage{amsmath}
\usepackage{enumitem}
\usepackage{tabularx}
\usepackage{multirow}
\usepackage{amsfonts}
\usepackage{float}
\usepackage{bm}
\usepackage{textcomp}
\usepackage{xcolor}

\begin{document}

\title{PainNet: Statistical Relation Network with Episode-Based Training for Pain Estimation\\
}

\author{\parbox{16cm}{\centering
    {Mina Bishay, Graham Page and Mohammad Mavadati}\\
    {\textit{Smart Eye AB}\\}
    {\tt\small \{mina.bishay, graham.page, mohammad.mavadati\}@smarteye.ai}
    }
}



\maketitle

\begin{abstract}

Despite the span in estimating pain from facial expressions, limited works have focused on estimating the sequence-level pain, which is reported by patients and used commonly in clinics. In this paper, we introduce a novel Statistical Relation Network, referred to as PainNet, designed for the estimation of the sequence-level pain. PainNet employs two key modules, the embedding and the relation modules, for comparing  pairs of pain videos, and producing relation scores indicating if each pair belongs to the same pain category or not. At the core of the embedding module is a statistical layer mounted on the top of a RNN for extracting compact video-level features. The statistical layer is implemented as part of the deep architecture. Doing so, allows combining multiple training stages used in previous research, into a single end-to-end training stage. PainNet is trained using the episode-based training scheme, which involves comparing a query video with a set of videos representing the different pain categories. Experimental results show the benefit of using the statistical layer and the episode-based training in the proposed model. Furthermore, PainNet outperforms the state-of-the-art results on self-reported pain estimation.




\end{abstract}


\section{INTRODUCTION}

Pain is characterized as an unpleasant sensation and distressing experience that is caused by potential tissue damage (like injuries) or numerous disorders \cite{ilana1979pain, williams2016updating}. Assessing the severity of pain is a crucial input for monitoring symptom progression and planning treatment. The predominant standard for pain assessment involves the subjective self-report provided by the patients, which is completed through different rating scales like the Visual Analog Scale (VAS) \cite{aicher2012pain, hawker2011measures}. Despite its common use, self-reported pain faces challenges in acquisition from certain patient groups, including toddlers, unconscious individuals, and those with severe mental disorders. Consequently, over the past decade, there has been a growing interest in developing automatic techniques for estimating pain intensity.

Facial expressions can serve as pivotal indicators of pain \cite{leresche1982facial, craig1992facial, prkachin2009assessing}, prompting numerous studies to concentrate on utilizing facial expressions or other facial features, such as landmarks, for pain estimation \cite{hassan2019automatic}. These endeavors typically align with two primary approaches; frame-level and sequence-level pain estimation. The first approach focuses on estimating pain frame-by-frame based on the intensity of some facial AUs, an example of this approach is the Prkachin and Solomon Pain Intensity (PSPI) metric \cite{prkachin2008structure}. The challenge with this approach lies in the subjective nature of pain, which is expressed differently across individuals due to factors like gender and age. As a result, it may not always be accurately reflecting the true pain. In contrast, sequence-level approaches like ours aim to estimate self-reported pain for each session (e.g., the VAS score) by capturing unique temporal characteristics throughout the entire session. This approach is commonly used in clinical settings. Despite this, a significant portion of the research has focused on predicting the frame-level PSPI scores, with limited attention given to the sequence-level pain estimation. This study aims to estimate the self-reported VAS score on the sequence level.

Inspired by the success of the relation networks in data-limited scenarios (i.e. few-shot learning) \cite{sung2018learning, bishay2019tarn}, and given that the pain dataset is relatively limited in data \cite{lucey2011painful}, we propose a statistical relation network (named PainNet) for estimating self-reported pain. PainNet is designed to compare various pairs of pain videos, and give as output relation scores that indicate whether each pair belongs to the same pain level or not. This involves two key modules: the embedding module and the relation module. In the \textbf{embedding module}, we extract a compact feature vector representing the whole video in 3 stages; first analyzing patients' facial expressions using the AFFDEX 2.0 \cite{bishay2023affdex} toolkit, then learning the temporal dynamics of the patients' expressions using a Gated Recurrent Unit (GRU) applied on short clips, and finally summarizing the clip-level features using a statistical layer. In the \textbf{relation module}, the extracted video-level embeddings are compared using metric learning, and then the relation scores are produced for the compared pairs of videos. 


In order to leverage the sequence-level pain estimation, we train our architecture using the episode-based training approach proposed in \cite{vinyals2016matching, snell2017prototypical}, where for each training episode, we randomly select a query example and a number of support examples for each of the detected classes. The query example is compared to the support examples through the relation network to find the best matching class. Episode-based training helps in minimizing inter-class variations and maximizing the intra-class variations \cite{hayale2019facial}. In few-shot learning, models learn general features by changing classes across episodes, while in pain estimation we have specific pain classes/levels but with more examples, so the support classes are kept fixed while changing the training examples across episodes. PainNet is trained in an end-to-end fashion using a weighted version of the binary-cross entropy function. Experimental results show that the episode-based training outperforms the conventional batch training, in addition we achieve state-of-the-art results on self-reported pain estimation.

The rest of the paper is organized as follows: we first review the related literature in self-reported pain estimation in Section 2. We then describe the proposed PainNet in Section 3. Finally, we give the experimental results and conclusion in Section 4 and Section 5, respectively. 


\section{Related Work}

Most of the work in automatic pain estimation has focused on using the Facial Action Coding System (FACS) \cite{EkmanBook97}, and specifically the AUs intensity for estimating the pain level frame-by-frame on the PSPI scale \cite{prkachin2008structure}, while relatively limited works have addressed the estimation of the self-reported pain on the sequence level using the VAS metric \cite{hassan2019automatic}. In this section, we focus on reviewing the works that have estimated the VAS pain score \cite {liu2017deepfacelift, lopez2017personalized, szczapa2021automatic, xu2019pain, erekat2020enforcing, xu2020exploring}, as it is the main goal of our paper. We review the existing works in terms of the model input, proposed methods, and training scheme.  


\textbf{Model input.} Different inputs have been used as facial features for pain estimation in the literature. In \cite{liu2017deepfacelift, lopez2017personalized, szczapa2021automatic}, the precise 66 facial landmarks provided with the UNBC dataset were used as input to the proposed models. In \cite {xu2019pain, erekat2020enforcing}, researchers used the relatively high-dimensional raw face images as input to a pre-trained Convolutional Neural Network (CNN) like AlexNet or VGGFace. In \cite{xu2020exploring}, Xu \textit{et al.} used the manually labeled AUs in the UNBC dataset as input for the proposed model, however, manual labeling is a hard and time-consuming process. In this paper, we use a low-dimensional feature vector consisting of the automatic predictions of 20 AUs as input to the proposed PainNet. 


\textbf{Proposed architectures.} Several works proposed methods that consists of multiple stages. First, some intermediate frame-level pain estimations (like PSPI scores) were extracted, and then those intermediate features were used for predicting the final sequence-level VAS score. Specifically, \cite{lopez2017personalized} proposed a two-step learning approach for pain estimation; first a RNN was used to estimate PSPI score on the frame-level, and then the estimated PSPI scores were passed to a personalized Hidden Conditional Random Fields to estimate the VAS score. In \cite{liu2017deepfacelift}, Liu \textit{et al.} proposed a two-stage personalized model (named DeepFaceLIFT) for pain estimation. In the first stage, a weakly-supervised MLP was trained for estimating the frame-level VAS scores, and then some sequence-level statistics were extracted from the outputs of the first stage and passed to a Gaussian process regression model for estimating the final VAS score. In \cite{xu2019pain}, Xu \textit{et al.} proposed a three-stage model for pain estimation; first a VGGFace is used to estimate PSPI scores on the frame-level. Then, 9 statistics were extracted over PSPI predictions and fed to a MLP for pain estimation. Finally, an optimal linear combination of the multidimensional sequence-level pain scores was used to predict the final VAS score.


In some other works, the sequence-level pain score was estimated directly using features extracted from the whole sequence. That is, \cite{xu2020exploring} extracted some statistical features based on the AUs and PSPI annotations of the whole sequence, and then those features are passed to an MLP for pain estimation. In \cite{erekat2020enforcing}, Erekat \textit{et al.} proposed an end-to-end spatio-temporal CNN-RNN model consisting of AlexNet and GRU for automatically estimating the VAS score. In our proposed model, the pain score is estimated directly using; a) a GRU that uses the AU predictions for extracting temporal features across short clips, and b) a statistical layer that summarizes the extracted GRU features. On the contrary to other works that have extracted some statistical features prior to pain estimation, our statistical layer has been implemented as a part of the deep learning model. 


\textbf{Training methodology.} The majority of studies in existing literature have utilized the UNBC dataset for training and testing, given that other datasets like EmoPain \cite{aung2015automatic}, SenseEmotionA \cite{velana2017senseemotion}, and X-ITE pain \cite{gruss2019multi} either offer solely frame-level pain annotations or were not publicly accessible  \cite{werner2019automatic}. 
The works that have extracted intermediate frame-level scores, were trained in multiple stages \cite{lopez2017personalized, liu2017deepfacelift, xu2019pain}, while the models proposed in \cite{xu2020exploring, erekat2020enforcing} were trained just once using the sequence-level VAS scores. Having multiple training stages is quite challenging in such data-limited scenario. Note that all the existing works have trained their models using the conventional batch training approach. In this paper, we train our model in a single stage using only the VAS labels, in addition, we propose to train our model using the episode-based training scheme that is commonly used in the few-shot learning problem.


\begin{figure*} [!t]
  \centering{\includegraphics[width=1.0 \linewidth]{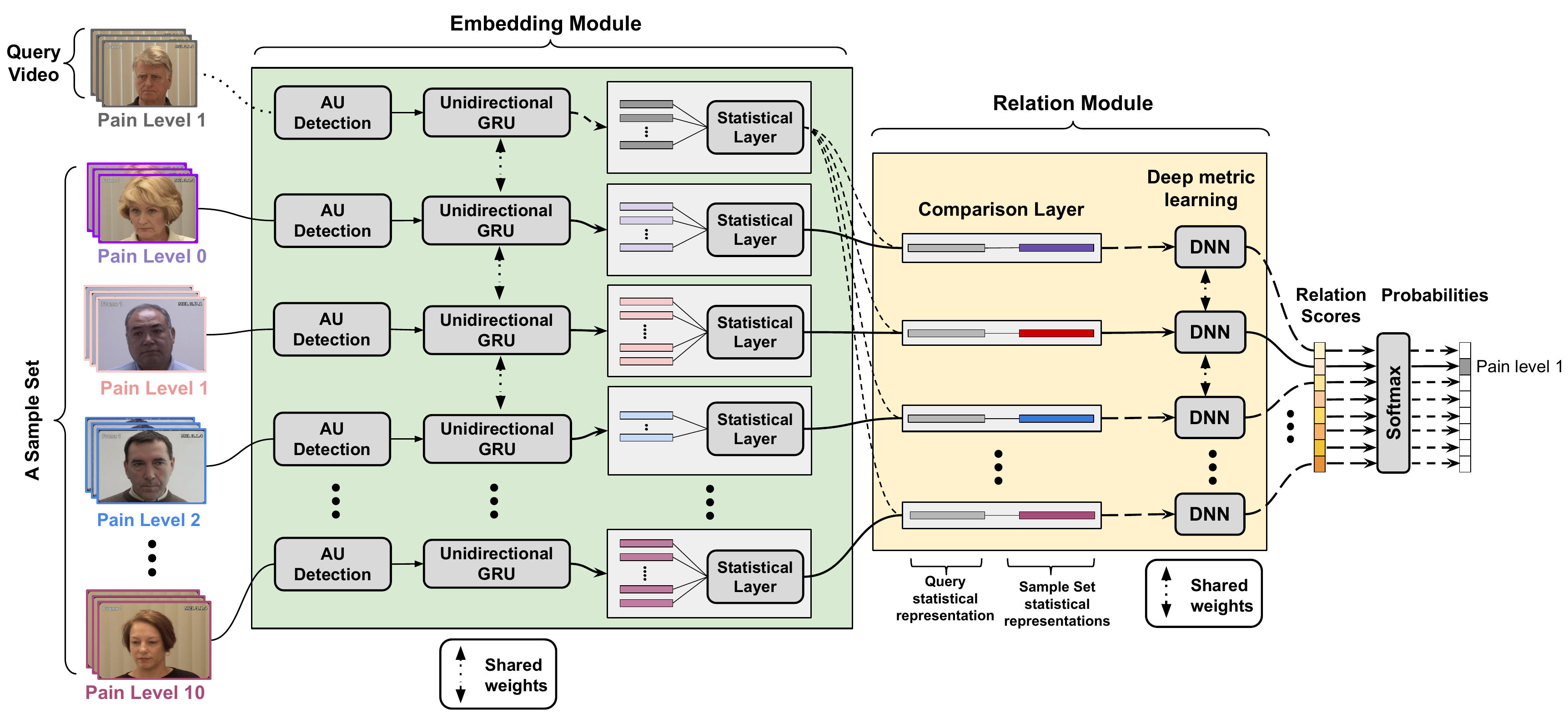}}
  \caption{The proposed architecture for estimating the sequence-level pain score.}
  \label{painet_method}
\end{figure*}


\section{Proposed Method}

In this section we introduce a novel architecture, named PainNet, for solving the problem of self-reported (VAS) pain estimation. Figure \ref{painet_method} shows an overview of the proposed method. PainNet learns through episode-based training to compare a query video to a sample set of videos representing the VAS pain levels, in order to find the best match for the query. Note that the sample set has 11 videos as the VAS score ranges between 0-10 (0 = low pain and 10 = severe pain). PainNet takes as input facial representations (i.e. AU predictions) extracted across the query and sample videos, and gives as output a relation score for each pair of the compared videos. More specifically, PainNet consists of two modules; the embedding module and the relation module. The embedding module processes the facial representations, first temporally on the segment-level and then statistically on the video-level, and produces a compact embedding for each video. The relation module compares the extracted video-level embeddings, and produces a matching score representing if the compared query and sample videos are coming from the same pain category or not.  These modules as well as the training approach are explained in detail in the following subsections.



\subsection{Embedding Module}

The embedding module consists of 3 main stages. We first analyze patients facial expressions using the AFFDEX 2.0 toolkit \cite{bishay2023affdex}. Then, we extract segment-wise temporal representations across the AU predictions using a GRU \cite{cho2014learning}. Finally, we calculate video-level statistical features across the temporal GRU features. 

\subsubsection{Facial Expression Analysis} 

particularly AU detection, is a crucial processing stage in numerous healthcare applications \cite{girard2014nonverbal, jaiswal2017automatic, bishay2019schinet, bishay2019can}. AUs serve as fundamental components of various facial expressions, and exhibit universality across diverse individuals. Recently, Bishay \textit{et al.} have introduced a new facial expression analysis toolkit, named AFFDEX 2.0 \cite{bishay2023affdex}, that can detect different AUs and emotions. AFFDEX 2.0 has been trained and tested using thousands of videos (circa 55,000 videos), that have been collected in-the-wild and from different demographic groups. AFFDEX 2.0 has achieved state-of-the-art results in AU detection and emotion recognition. 


In our analysis, we use AFFDEX 2.0 for predicting the occurrence probability of 20 different AUs across each video frame.  We center the AU predictions for each video by subtracting the mean over each AU from the AU predictions in the whole video. The predictions for each AU are centered as follows:
\begin{equation}
\begin{array}{l}
AU_{mean} = \dfrac{1}{T} \sum_{t=1}^{T} AU_{t} \\
AU_{t} = (AU_{t} - AU_{mean}), \quad t \in \{1,2,..., T\} \\
\end{array}
\end{equation}
where $T$ represents the total number of frames or timestamps in the video. The centered AU predictions act as the first level of feature extraction in the embedding module (i.e. frame-level features), and are used as input to the GRU model. List of the AUs used in our analysis are shown in Table \ref{table_AUs_1}.



\begin{table*}[!t]
\caption{List of the AUs/expressions predicted using the AFFDEX 2.0 toolkit \cite{bishay2023affdex}.}
\label{table_AUs_1}
\begin{adjustbox}{width=1.0 \textwidth, center}
\begin{threeparttable}
  \centering
\begin{tabular}{|c||c|c|c|c|c|c|c|c|c|c|c|c|c|c|c|c|c|c|c|c|}
\hline

Action Unit (AU) & AU1 & AU2 & AU4 & AU5 & AU6 & AU7 & AU9 & AU10 & AU12 & AU14 & AU15 & AU17 & AU18 & AU20 & AU24 & AU25 & AU26 & AU28 & AU43 &  -  \\ \hline \hline 

\textbf{\ \ \parbox{2cm}{ \quad \ Facial \\ Expressions}}  & 
\textbf{\rotatebox[origin=c]{90}{\parbox{2cm}{Inner Brow \\ Raiser}}} & 
\textbf{\rotatebox[origin=c]{90}{\parbox{2cm}{Outer Brow \\ Raiser}}} & 
\textbf{\rotatebox[origin=c]{90}{\parbox{2cm}{Brow Furrow}}} & 
\textbf{\rotatebox[origin=c]{90}{\parbox{2cm}{Upper Lid \\ Raiser}}} & 
\textbf{\rotatebox[origin=c]{90}{\parbox{2cm}{Cheek Raiser}}} &
\textbf{\rotatebox[origin=c]{90}{\parbox{2cm}{Lid Tightener}}} &
\textbf{\rotatebox[origin=c]{90}{\parbox{2cm}{Nose Wrinkle}}} &
\textbf{\rotatebox[origin=c]{90}{\parbox{2cm}{Upper Lip \\ Raiser}}} &
\textbf{\rotatebox[origin=c]{90}{\parbox{2cm}{Lip Corner \\ Puller}}} &
\textbf{\rotatebox[origin=c]{90}{\parbox{2cm}{Dimpler}}} &
\textbf{\rotatebox[origin=c]{90}{\parbox{2cm}{Lip Corner \\ Depressor}}} &
\textbf{\rotatebox[origin=c]{90}{\parbox{2cm}{Chin Raiser}}} &
\textbf{\rotatebox[origin=c]{90}{\parbox{2cm}{Lip Pucker}}} &
\textbf{\rotatebox[origin=c]{90}{\parbox{2cm}{Lip Stretch}}} &
\textbf{\rotatebox[origin=c]{90}{\parbox{2cm}{Lip Press}}} &
\textbf{\rotatebox[origin=c]{90}{\parbox{2cm}{Lips Part}}}   & 
\textbf{\rotatebox[origin=c]{90}{\parbox{2cm}{Jaw Drop}}}   & 
\textbf{\rotatebox[origin=c]{90}{\parbox{2cm}{Lip Suck}}}   & 
\textbf{\rotatebox[origin=c]{90}{\parbox{2cm}{Eyes Closure}}}   & 
\textbf{\rotatebox[origin=c]{90}{\parbox{2cm}{Smirk}}}  \\ \hline

 \end{tabular}
  \begin{tablenotes}
\item Smirk is defined as the asymmetric lip corner puller (AU12) or dimpler (AU14).
\end{tablenotes}
\end{threeparttable}
\end{adjustbox}
\end{table*}


\subsubsection{Temporal modeling}

The AU predictions extracted from the query and sample videos are temporally divided into short segments, each consisting of 16 frames (around 0.5 second). No overlap or downsampling is applied during segmentation. The AU predictions from each segment are passed to a uni-directional GRU for extracting a short representation at the last time step. Figure \ref{painet_method} shows the different representations/embeddings extracted from the query and sample videos. Note that the number of representations produced by the GRU depends on the length of the processed video. The GRU consists of a single hidden layer of size $d$. Dropout is used for GRU regularization. 

\subsubsection{Statistical layer}

Several works in the literature have used intermediate frame-level pain scores/representations for calculating a number of statistics, that are then passed to another regression model for estimating the final pain score \cite{liu2017deepfacelift, xu2019pain, xu2020exploring}. On the contrary, we implement a custom layer in our deep architecture for statistics calculation (referred to as statistical layer), and subsequently combine the two training stages used in previous works into a single end-to-end training stage.  


The statistical layer consists of 4 neurons, each of which is dedicated for calculating one of the following 4 statistics; mean, median, Standard Deviation (Std) and LogSumExp (LSE). The statistics calculation is conducted on GRU neuron basis, that is, the statistics are calculated separately for each GRU hidden unit. Given a query video $\bm{Q} \in \mathbb{R}^{M \times d}$, where each row in $\bm{Q}$ represents a segment-embedding vector of dimension $d$, and where $M$ denotes the number of segments in the video $\bm{Q}$. The statistical layer gives as output matrix $\bm{S} \in \mathbb{R}^{4 \times d}$, where each row in $\bm{S}$ represents the output for one of the 4 statistical operators. Note that $\bm{S}$ does not depend on the number of video segments. The 4 statistics are calculated as follows: 
\begin{equation}
S = 
\begin{dcases}
S_{mean_i} = \dfrac{1}{M} \sum_{m=1}^{M} Q_{mi} \\
S_{med_i} = median(Q_{1i}, Q_{2i}, ..., Q_{Mi}) \\
S_{Std_i} = \sqrt{\dfrac{1}{M} \sum_{m=1}^{M} (Q_{mi} - S_{mean_i})^2 } \\
S_{LSE_i} = log \sum_{m=1}^{M} e^{Q_{mi}} \\
\end{dcases}
\end{equation}
where $i$ represents a GRU neuron and has values ranging between $1$ and $d$.
 

The output of the statistical layer, $\bm{S}$, is then flattened in order to have a single vector representing the query video. The extracted vector has a dimensionality of $1 \times 4d$. Similarly, a set of vectors are extracted to represent the sample set videos. The statistics calculated in the statistical layer are predefined to reduce complexity, and subsequently the statistical layer has no learnable parameters. Batch normalization is used after each layer in the embedding module for better training convergence.

\subsection{Relation Module}

In the relation module, we pair the video-level embedding extracted from the query video with each embedding extracted from the sample set, and then we compare the embeddings and produce a single relation score for each pair. The relation module consists of a comparison layer and a non-linear classifier for performing metric learning. The comparison layer calculates a distance/similarity measure between the video-level features extracted from each video pair. The comparison layer outputs are then passed to a network that gives as output a relation score representing if the two compared videos belong to the same pain category or not. Finally, the relation scores produced from comparing the query video to all the sample set videos are passed to a softmax layer for mapping the relation scores to a probability distribution over the different pain levels. 


According to \cite{wang_17_match}, the comparison layer can be one of the following operations: multiplication (Mult), subtraction (Subt), neural network (NN), subtraction and multiplication followed by a neural network (SubMultNN), or Euclidean distance and cosine similarity (EucCos). In the experimental section, we will show that the EucCos operation performs the best in pain level estimation. The network following the comparison layer consists of two fully-connected layers with a ReLU activation. 

\subsection{Training scheme}

In this section, we first explain how we adapt the episode-based training for the pain estimation problem. Second, we clarify the loss function used for training our architecture. 

\subsubsection{Episode-Based Training}

To the best of our knowledge, all the previous works in pain estimation have trained their architectures using batch training. Episode-based training is another training scheme that has been used frequently in the problem of few-shot learning \cite{vinyals_16_oneshot, snell_17_prototypical, bishay2019tarn}, where a number of episodes are formed during the training phase based on a large related dataset. During each episode, we randomly choose a subset of classes from the training set, and for those classes we select  a query example and a number of examples within each class to act as the sample set. Then, the query example is compared to each example in the sample set to find the best matching class, and subsequently classifying the query. According to \cite{hayale2019facial}, using episode-based training with relation/Siamese networks helps in reducing the inter-class variations and increasing the intra-class variations between the compared classes. So, in this paper we propose to use episode-based training for pain level estimation. 


Unlike the few-shot learning problem where a few training examples (1-5 examples/class) are available for the intended classes, and a large dataset for other classes related to the intended ones, in the pain level estimation problem we have relatively more training examples for the intended classes (on average $\sim$18 examples/class), but with no large related dataset. Subsequently, we form the training and testing episodes in pain level estimation from just the intended classes (i.e. pain categories).



In each training episode, we randomly select a query video from the training set, and a sample set of videos representing the pain categories in the VAS rating, that is, a single video per each pain category forming by that a sample set consisting of 11 videos. During validation and testing, the query video is selected from the validation or the testing sets, while the sample set is selected from the training set. In few-shot learning, sample sizes typically range from 1 to 5, with larger sample sets generally resulting in improved performance. However, due to limitations in available samples for certain pain levels, we opt to use 5 sample sets during validation and testing. The final classification score is determined by calculating the median across the model's outputs. Both the embedding and the relation modules (excluding the AFFDEX 2.0 toolkit) are trained in an end-to-end fashion.

%






\subsubsection{Loss function}

The relation networks are used for classification problems, giving as output a set of binary labels indicating if the query matches the sample set videos or not, so the Binary Cross-Entropy (BCE) was commonly used in previous works as a cost function. However, pain level estimation is an ordinal regression problem, so the error between the different pain categories should not be treated equally. Subsequently, we use an adapted version of the BCE function for training our network. That is, we multiply BCE loss with the absolute difference between the predicted and the ground truth pain scores. We name this cost function the Weighted Binary Cross-Entropy ($W_{BCE}$), and for each episode the cost is calculated as follows:
\begin{equation}
W_{BCE} = \dfrac{1}{C} \sum_{c=1}^{C} ((|T - c| + 1) \cdot BCE_{c}) 
\end{equation}
\begin{equation}
BCE_{c} = - (y_{c} \log(p_{c}) + (1-y_{c}) \log(1-p_{c}))
\end{equation}
where $T$ denotes the pain ground truth label, $C$ the number of classes represented in the support set (VAS has 11 pain labels), $y_{c}$ is a binary label representing if the query video matches the sample video at class $c$ or not, $p_{c}$ is the network output probability at class $c$. Note that we add a value of 1 to the absolute difference in equation 3 to avoid multiplication by zero. 






\section{Experimental Results}

\subsection{Dataset}

For our experiments, we use the UNBC-McMaster dataset \cite{lucey2011painful}, which is widely used for pain level estimation. It consists of 200 videos recorded for 25 patients, who are suffering from shoulder pain. Patients were recorded while performing some shoulder exercises. Each video in the UNBC dataset is annotated by 4 sequence-level pain labels; VAS (Visual Analog Scale), OPR (Observers Pain Rating), AFF (Affective-motivational scale) and SEN (Sensory Scale). The first three ratings are reported by the patients themselves, while the OPR is an observer pain rating. The dataset was also manually labeled in terms of the intensity of 11 AUs. Our goal is to estimate the self-reported VAS score. 


Similar to other works in the literature \cite{liu2017deepfacelift, xu2019pain, xu2020exploring, szczapa2021automatic, szczapa2022automatic}, we partition the UNBC dataset into 5 folds, where each data fold consists of the videos of 5 patients. Then, we iteratively combine 4 folds for training and use 1 fold for testing. During each trial, we randomly select 10 videos from the training set for validation. The average performance across the 5 folds is reported in our experiments.


\subsection{Performance metrics}

Following \cite{xu2019pain, xu2020exploring}, we use three metrics for reporting the performance of pain level estimation; the Intra-Class Correlation (ICC) \cite{shrout1979intraclass}, the Mean Absolute Error (MAE), and the Root Mean Square Error (RMSE). MAE is a good indication of accuracy, however, it is less sensitive to outliers, and sometimes is falsely low (specifically when the model predicts everything close to the dataset mean). Subsequently, we use the RMSE to highlight large errors, and the ICC to measure how the predicted values vary/correlate with the ground truth values. In what follows we report the 3 metrics.

\subsection{Implementation details}

The AU predictions are embedded using a unidirectional GRU layer with a size of 16. The network used for metric learning has two fully-connected layers of size 2 and 1. The proposed architecture is trained for 1500 episodes, and the model weights are updated every 5 episodes. The AU predictions are augmented by random Gaussian noise to reduce the overfitting problem. The best-performing model on the validation set is used for testing. We train our architecture using the ADAM optimiser with a learning rate equal to $0.005$ and gradient clipping set to $1$. Dropout with a probability of $0.5$ is used for GRU regularization. We evaluate our architecture every 50 training episodes. We use the PyTorch library for implementing the PainNet architecture.

\subsection{Ablation Studies}

\subsubsection{Embedding module}

In our first experiment, we compare the proposed embedding module consisting of a unidirectional GRU and a statistical layer to the embedding modules used in the relation networks of \cite{bishay2019tarn} and \cite{bishay2020automatic}, which consists of two stacked GRUs; one extracting features across short video segments, while the other is summarizing the segment-level features, and giving as output a single embedding representing the whole video. The stacked GRUs tested here consist of a single hidden layer, whose size is selected according to the results on the validation set. Table~\ref{training_table} shows the performance across the two embedding modules. Results show that using the statistical layer for summarizing the segment-level features has better performance than using another GRU. On average, ICC is improved by around 8\%, and similar improvement can be seen in MAE and RMSE.


\subsubsection{Training strategy}

In the next experiment, we compare the conventional batch training to the episode-based training across the two embedding modules, explained in the previous experiment. In the batch training, the training set videos are divided randomly into a number of batches, and the error calculated across those batches is used for updating the model weights. The batch size is selected according to the results on the validation set. Table~\ref{training_table} shows the performance across the two training strategies. Results show that using episode-based training has better results than the batch training. On average, the episode-based training along with the proposed embedding module has improved ICC by $\sim$12\% and RMSE by $\sim$20\%.  


\begin{table}[!t]
\centering
\caption{The pain estimation performance of two embedding modules that are trained using two different strategies.}
\begin{adjustbox}{width=\columnwidth}
    \begin{tabular}{ | c | c || c | c | c |} \hline
    \textbf{Embedding Module}  & \textbf{Training type}  & \textbf{ICC}  & \textbf{MAE} & \textbf{RMSE} \\ \hline  \hline
    \multirow{2}{*}{Two Stacked GRUs} & Batch & 0.51  & 1.94  &  2.48  \\ \cline{2-5} 
                              & Episode & 0.59  & 1.92  & 2.40   \\ \hline
    \multirow{2}{*}{GRU + Statistical layer} & Batch & 0.57  & 1.93 &  2.69 \\ \cline{2-5} 
                              & Episode & \textbf{0.71}  & \textbf{1.53} &  \textbf{2.08} \\ \hline
    \end{tabular}
\end{adjustbox}
\label{training_table}
\end{table}


\subsubsection{Statistical layer}
In the third experiment, we investigate the impact of increasing gradually the size of the statistical layer to include more operators. Specifically, we first start by using a single statistical operator which is the mean, and then we gradually add the other 3 operators (standard deviation, LogSumExp, median), and in each case we measure the performance of our proposed model. Table~\ref{statistics_table} shows the performance across the different statistical operators. Results show that increasing the statistical layer size to include more operators has a positive impact on the performance. The improved performance can be seen across all metrics. Note that adding other operators like minimum and maximum has slightly degraded the performance.  


\begin{table}[!t]
\centering
\caption{Comparison between different statistical operators included in the statistical layer of the PainNet model.}
\begin{adjustbox}{width=\columnwidth}
    \begin{tabular}{ | c || c | c | c |} \hline
    \textbf{Statistical Layer Operators}  & \textbf{ICC}  & \textbf{MAE} & \textbf{RMSE} \\ \hline  \hline
    \textbf{Mean} &  0.53  & 1.85  &  2.45  \\ \hline 
    \textbf{Mean, Std} & 0.61  & 1.79  & 2.39   \\ \hline
    \textbf{Mean, Std, LSE} & 0.69  & 1.59 &  2.16 \\ \hline
    \textbf{Mean, Std, LSE, Median} &  \textbf{0.71}  & \textbf{1.53} &  \textbf{2.08} \\ \hline
    \textbf{Mean, Std, LSE, Median, Min, Max} & 0.66  & 1.58 &  2.26 \\ \hline
    \end{tabular}
\end{adjustbox}
\label{statistics_table}
\end{table}


\subsubsection{AU detection}

In the fourth experiment, we explore how the number of AUs utilized and the accuracy of AU detection impact pain estimation accuracy. Initially, we compare the performance of using the 10 AUs annotated in the UNBC dataset with the 20 AUs detected by AFFDEX 2.0 (detailed in Table I). Next, we replicate the initial experiment but substitute AFFDEX 2.0 with less effective AU detector, AFFDEX 1.0. Finally, we assess the impact of using the ground truth labels of the 10 AUs provided in the UNBC dataset instead of relying on AFFDEX 2.0. From Table \ref{sets_AUs_table}, we observe that employing a larger number of AUs enhances performance. Furthermore, comparing pain estimation results across AFFDEX 1.0, AFFDEX 2.0, and ground truth labels highlights that superior AU detection contributes to improved accuracy in pain estimation. Notably, AFFDEX 2.0 achieves performance levels closely aligned with those achieved using ground truth AU labels.

\begin{table}[!t]
\centering
\caption{Comparison between different AU predictions in the PainNet model.}
\begin{adjustbox}{width=0.8\columnwidth}
    \begin{tabular}{ | c || c | c | c |} \hline
    \textbf{No. of AUs (model used)}  & \textbf{ICC}  & \textbf{MAE} & \textbf{RMSE} \\ \hline  \hline

    \textbf{10 AUs (AFFDEX 1.0)} & 0.49  & 2.20 &  2.78   \\ \hline
    \textbf{20 AUs (AFFDEX 1.0)} & 0.56  & 2.03 &  2.68   \\ \hline \hline

    \textbf{10 AUs (Ground truth)} & 0.68  & \textbf{1.53} &  2.22   \\ \hline  \hline

    \textbf{10 AUs (AFFDEX 2.0)} & 0.67  & 1.65  & 2.28   \\ \hline
    \textbf{20 AUs (AFFDEX 2.0)} & \textbf{0.71}  & \textbf{1.53} &  \textbf{2.08}   \\ \hline
    
    
    \end{tabular}
\end{adjustbox}
\label{sets_AUs_table}
\end{table}


\subsubsection{Distance/Similarity measure}

In the fifth experiment, we investigate the effect of the different distance/similarity functions that can be used in the comparison layer, on the PainNet performance. \cite{wang_17_match} has compared five different similarity functions (Mult, Subt, NN, SubMultNN, EucCos). Table \ref{Distance_table} shows the performance obtained by PainNet over the different functions. EucCos has the best ICC, MAE and RMSE across the 5 functions. 



\begin{table}[!t]
\centering
\caption{Comparison between different similarity/distance measures included in the comparison layer of the PainNet model.}
\begin{adjustbox}{width=0.75\columnwidth}
    \begin{tabular}{ | c || c | c | c |} \hline
    \textbf{Similarity measure}  & \textbf{ICC}  & \textbf{MAE} & \textbf{RMSE} \\ \hline  \hline     \textbf{NN} & 0.45  & 2.38 &  3.24 \\ \hline
    \textbf{Subt} & 0.57  & 1.91  & 2.67   \\ \hline
    \textbf{Mult} &  0.62  & 1.93  &  2.72  \\ \hline 
    \textbf{SubMultNN} & 0.56 & 1.95 &  2.82 \\ \hline
    \textbf{EucCos} & \textbf{0.71}  & \textbf{1.53} &  \textbf{2.08} \\ \hline
    \end{tabular}
\end{adjustbox}
\label{Distance_table}
\end{table}


\subsubsection{Loss function}

In the last experiment, we compare the impact of using the BCE loss function to the weighted BCE function, which takes into account that pain estimation is an ordinal regression problem. Table \ref{loss_table} shows the performance obtained by the PainNet model over the two loss functions. Weighted BCE has better performance than the BCE function.


\begin{table}[!t]
\centering
\caption{Comparison between using the BCE and the weighted-BCE loss functions for training the PainNet model.}
\begin{adjustbox}{width=0.67\columnwidth}
    \begin{tabular}{ | c || c | c | c |} \hline
    \textbf{Loss function}  & \textbf{ICC}  & \textbf{MAE} & \textbf{RMSE} \\ \hline  \hline
    \textbf{BCE} & 0.66  & 1.62  & 2.29   \\ \hline
    \textbf{WBCE} &  \textbf{0.71}  & \textbf{1.53} &  \textbf{2.08} \\ \hline
    \end{tabular}
\end{adjustbox}
\label{loss_table}
\end{table}


\begin{table*}[t]
\centering
\caption{Performance of the proposed PainNet model as well as other state-of-the-art methods on self-reported pain estimation.}
\begin{adjustbox}{width=0.8 \textwidth, center}
    \begin{tabular}{ | c || c | c || c | c | c |} \hline
    \textbf{Method}  & \textbf{Model Input}  &  \textbf{Training Labels} &  \textbf{ICC}  & \textbf{MAE} & \textbf{RMSE} \\ \hline  \hline
    \textbf{DeepFaceLift \cite{liu2017deepfacelift}} & Facial landmarks & VAS, OPR & 0.35&  2.18 &  -  \\ \hline 
    \textbf{Extended MTL from pixel \cite{xu2019pain}} & Face images & AUs, PSPI, VAS, OPR, AFF, SEN & 0.43 & 1.95 & 2.45  \\ \hline
    \textbf{CNN-RNN \cite{erekat2020enforcing}} & Face images & VAS, OPR, AFF, SEN & - & 2.34  & -  \\ \hline
    \textbf{Extended MTL with AUs \cite{xu2020exploring}} & AUs, PSPI &  VAS, OPR, AFF, SEN & 0.61 & 1.73 & 2.15 \\ \hline
    \textbf{Manifold trajectories \cite{szczapa2021automatic}} & Facial landmarks & VAS & - & 2.44 & 3.15  \\ \hline
    \textbf{Trajectory Analysis \cite{szczapa2022automatic}} & Facial landmarks  & VAS & - & 1.59 & \textbf{1.98}  \\ \hline \hline
    \textbf{PainNet} & AUs & VAS & \textbf{0.71}  & \textbf{1.53} &  2.08 \\ \hline
    \end{tabular}
\end{adjustbox}
\label{literature_table}
\end{table*}


\subsection{Comparison to state of the art in pain level estimation}

In this section, we compare PainNet to the models proposed in the literature for estimating the self-reported pain score. Those methods include DeepFaceLift \cite{liu2017deepfacelift}, Extended MTL from pixel \cite{xu2019pain}, CNN-RNN \cite{erekat2020enforcing}, Extended MTL with AUs \cite{xu2020exploring}, Manifold trajectories \cite{szczapa2021automatic}, and Trajectory Analysis \cite{szczapa2022automatic}. For all the previous methods, we report the best performance achieved by the proposed model, and for \cite{xu2020exploring} we exclude results that use the human pain labels as input to their model. We did not compare to the method proposed in \cite{lopez2017personalized}, as it has been tested using a different protocol. Table~\ref{literature_table} summarizes the performance across the different architectures. Our architecture outperforms all the other methods on ICC and MAE, and has slightly less RMSE than the method proposed in \cite{szczapa2022automatic}. Table~\ref{literature_table} also highlights for each method the type of facial features used as input, and the kind of pain labels used for model training.



\begin{figure}[!t]
  \centering{\includegraphics[width=0.99\linewidth]{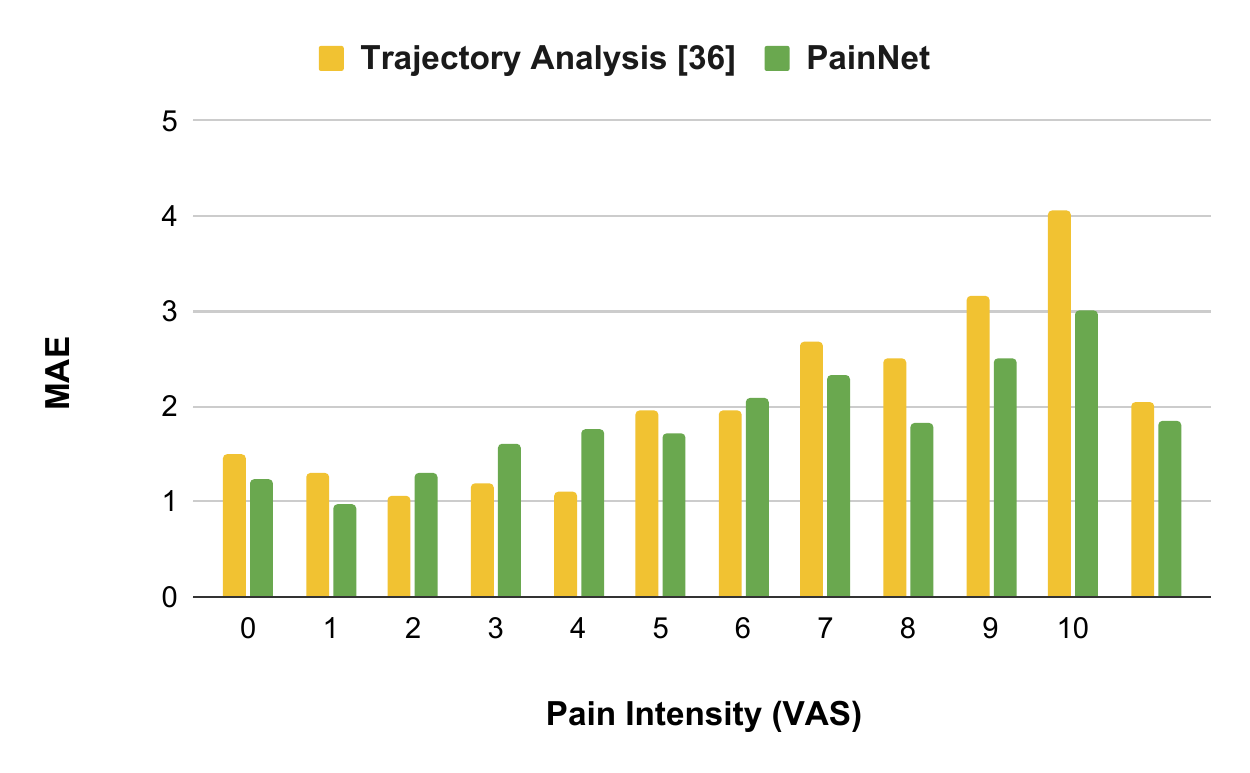}}
  \caption{MAE per intensity for the PainNet and Trajectory Analysis \cite{szczapa2022automatic} models.}
  \label{figure_MAE}
\end{figure}

In order to have a better comparison between PainNet and the method proposed in \cite{szczapa2022automatic}, we investigate the distribution of the MAE across the different pain intensities. Figure \ref{figure_MAE} shows the MAE per intensity for both methods. Although the two methods have a close performance on the MAE, we can see that PainNet has more balanced error across the different pain levels than \cite{szczapa2022automatic} -- this can be clearly seen across the high pain levels. Maintaining a balanced error is a crucial step when addressing problems with imbalanced data, such as pain estimation. This ensures that the trained model is not biased toward the most frequent classes (i.e. pain levels with more sessions). Furthermore, PainNet improves the average MAE/intensity by around 10\% compared to \cite{szczapa2022automatic} (On average MAE/intensity is 1.84 for PainNet vs 2.03 for \cite{szczapa2022automatic}).


\section{Conclusions and Future work}

Although numerous papers focus on estimating pain from facial expressions, few have concentrated on evaluating the sequence-level pain, which is reported by patients and commonly used in clinical settings. To address this gap, we propose a statistical relation network, named PainNet, for self-reported pain estimation. PainNet learns to compare a query video to a sample set of videos representing the whole pain classes in order to find the best matching class. 

PainNet consists of two modules; the embedding module which encodes the pain videos into short embeddings by using a unidirectional GRU and a statistical layer on top of it, and the relation module which compares the different embeddings by using metric learning. Unlike other works, the statistical layer is implemented as part of the deep learning architecture, allowing our model to get trained in a single end-to-end stage. In addition, our model is trained using episode-based training instead of conventional batch training. 

Our experimental results show that using the proposed statistical layer and episode-based training improves the performance of pain level estimation. Furthermore, PainNet outperforms other methods in the literature, and achieves the best results in ICC and MAE. In our future work, we plan to include additional modalities such as speech and body gestures alongside facial expressions. We also aim to evaluate ratings from external observers.

\section*{Ethical Impact Statement}


Our study addresses a practical application: estimating self-reported pain, a crucial measure in clinical settings. We utilize the publicly available UNBC dataset for both training and testing. In terms of model generalisability, the underlying SDK for AU detection has been trained using videos from participants recruited globally, suggesting robust performance across diverse demographics. However, our pain estimation model is specifically trained on the UNBC dataset, which has limited subject diversity. To have a more generalizable automated pain estimation model, we advocate for collaborative efforts between academia and industry to gather larger and more diverse clinical pain datasets, aiming to enhance the development of more universally applicable models.


\section*{Acknowledgment}

We would like to highlight that this work is not dedicated for any commercial product, and is only dedicated for research purposes. 

{\small
\bibliographystyle{IEEEtran}
\bibliography{egbib}

\begin{thebibliography}{10}
\providecommand{\url}[1]{#1}
\csname url@samestyle\endcsname
\providecommand{\newblock}{\relax}
\providecommand{\bibinfo}[2]{#2}
\providecommand{\BIBentrySTDinterwordspacing}{\spaceskip=0pt\relax}
\providecommand{\BIBentryALTinterwordstretchfactor}{4}
\providecommand{\BIBentryALTinterwordspacing}{\spaceskip=\fontdimen2\font plus
\BIBentryALTinterwordstretchfactor\fontdimen3\font minus \fontdimen4\font\relax}
\providecommand{\BIBforeignlanguage}[2]{{%
\expandafter\ifx\csname l@#1\endcsname\relax
\typeout{** WARNING: IEEEtran.bst: No hyphenation pattern has been}%
\typeout{** loaded for the language `#1'. Using the pattern for}%
\typeout{** the default language instead.}%
\else
\language=\csname l@#1\endcsname
\fi
#2}}
\providecommand{\BIBdecl}{\relax}
\BIBdecl

\bibitem{ilana1979pain}
E.~Ilana, ``Pain terms; a list with definitions and notes on usage,'' \emph{Pain}, vol.~6, p. 249, 1979.

\bibitem{williams2016updating}
A.~C. d.~C. Williams and K.~D. Craig, ``Updating the definition of pain,'' \emph{Pain}, vol. 157, no.~11, pp. 2420--2423, 2016.

\bibitem{aicher2012pain}
B.~Aicher, H.~Peil, B.~Peil, and H.~Diener, ``Pain measurement: Visual analogue scale (vas) and verbal rating scale (vrs) in clinical trials with otc analgesics in headache,'' \emph{Cephalalgia}, vol.~32, no.~3, pp. 185--197, 2012.

\bibitem{hawker2011measures}
G.~A. Hawker, S.~Mian, T.~Kendzerska, and M.~French, ``Measures of adult pain: Visual analog scale for pain (vas pain), numeric rating scale for pain (nrs pain), mcgill pain questionnaire (mpq), short-form mcgill pain questionnaire (sf-mpq), chronic pain grade scale (cpgs), short form-36 bodily pain scale (sf-36 bps), and measure of intermittent and constant osteoarthritis pain (icoap),'' \emph{Arthritis care \& research}, vol.~63, no. S11, pp. S240--S252, 2011.

\bibitem{leresche1982facial}
L.~LeResche, ``Facial expression in pain: A study of candid photographs,'' \emph{Journal of Nonverbal Behavior}, vol.~7, pp. 46--56, 1982.

\bibitem{craig1992facial}
K.~D. Craig, K.~M. Prkachin, and R.~V. Grunau, ``The facial expression of pain,'' \emph{Handbook of pain assessment}, vol.~2, pp. 257--276, 1992.

\bibitem{prkachin2009assessing}
K.~M. Prkachin \emph{et~al.}, ``Assessing pain by facial expression: facial expression as nexus,'' \emph{Pain Research and Management}, vol.~14, pp. 53--58, 2009.

\bibitem{hassan2019automatic}
T.~Hassan, D.~Seu{\ss}, J.~Wollenberg, K.~Weitz, M.~Kunz, S.~Lautenbacher, J.-U. Garbas, and U.~Schmid, ``Automatic detection of pain from facial expressions: a survey,'' \emph{IEEE transactions on pattern analysis and machine intelligence}, vol.~43, no.~6, pp. 1815--1831, 2019.

\bibitem{prkachin2008structure}
K.~M. Prkachin and P.~E. Solomon, ``The structure, reliability and validity of pain expression: Evidence from patients with shoulder pain,'' \emph{Pain}, vol. 139, no.~2, pp. 267--274, 2008.

\bibitem{sung2018learning}
F.~Sung, Y.~Yang, L.~Zhang, T.~Xiang, P.~H. Torr, and T.~M. Hospedales, ``Learning to compare: Relation network for few-shot learning,'' in \emph{Proceedings of the IEEE conference on computer vision and pattern recognition}, 2018, pp. 1199--1208.

\bibitem{bishay2019tarn}
M.~Bishay, G.~Zoumpourlis, and I.~Patras, ``Tarn: Temporal attentive relation network for few-shot and zero-shot action recognition,'' \emph{arXiv preprint arXiv:1907.09021}, 2019.

\bibitem{lucey2011painful}
P.~Lucey, J.~F. Cohn, K.~M. Prkachin, P.~E. Solomon, and I.~Matthews, ``Painful data: The unbc-mcmaster shoulder pain expression archive database,'' in \emph{2011 IEEE International Conference on Automatic Face \& Gesture Recognition (FG)}.\hskip 1em plus 0.5em minus 0.4em\relax IEEE, 2011, pp. 57--64.

\bibitem{bishay2023affdex}
M.~Bishay, K.~Preston, M.~Strafuss, G.~Page, J.~Turcot, and M.~Mavadati, ``Affdex 2.0: A real-time facial expression analysis toolkit,'' in \emph{2023 IEEE 17th International Conference on Automatic Face and Gesture Recognition (FG)}.\hskip 1em plus 0.5em minus 0.4em\relax IEEE, 2023, pp. 1--8.

\bibitem{vinyals2016matching}
O.~Vinyals, C.~Blundell, T.~Lillicrap, D.~Wierstra \emph{et~al.}, ``Matching networks for one shot learning,'' \emph{Advances in neural information processing systems}, vol.~29, 2016.

\bibitem{snell2017prototypical}
J.~Snell, K.~Swersky, and R.~Zemel, ``Prototypical networks for few-shot learning,'' \emph{Advances in neural information processing systems}, vol.~30, 2017.

\bibitem{hayale2019facial}
W.~Hayale, P.~Negi, and M.~Mahoor, ``Facial expression recognition using deep siamese neural networks with a supervised loss function,'' in \emph{2019 14th IEEE International Conference on Automatic Face \& Gesture Recognition (FG 2019)}.\hskip 1em plus 0.5em minus 0.4em\relax IEEE, 2019, pp. 1--7.

\bibitem{EkmanBook97}
P.~Ekman and E.~L. Rosenberg, \emph{What the face reveals: Basic and applied studies of spontaneous expression using the Facial Action Coding System (FACS)}.\hskip 1em plus 0.5em minus 0.4em\relax Oxford University Press, USA, 1997.

\bibitem{liu2017deepfacelift}
D.~Liu, P.~Fengjiao, R.~Picard \emph{et~al.}, ``Deepfacelift: interpretable personalized models for automatic estimation of self-reported pain,'' in \emph{IJCAI 2017 Workshop on Artificial Intelligence in Affective Computing}.\hskip 1em plus 0.5em minus 0.4em\relax PMLR, 2017, pp. 1--16.

\bibitem{lopez2017personalized}
D.~Lopez~Martinez, R.~Picard \emph{et~al.}, ``Personalized automatic estimation of self-reported pain intensity from facial expressions,'' in \emph{Proceedings of the IEEE conference on computer vision and pattern recognition workshops}, 2017, pp. 70--79.

\bibitem{szczapa2021automatic}
B.~Szczapa, M.~Daoudi, S.~Berretti, P.~Pala, A.~Del~Bimbo, and Z.~Hammal, ``Automatic estimation of self-reported pain by interpretable representations of motion dynamics,'' in \emph{2020 25th International Conference on Pattern Recognition (ICPR)}.\hskip 1em plus 0.5em minus 0.4em\relax IEEE, 2021, pp. 2544--2550.

\bibitem{xu2019pain}
X.~Xu, J.~S. Huang, and V.~R. De~Sa, ``Pain evaluation in video using extended multitask learning from multidimensional measurements.'' in \emph{ML4H@ NeurIPS}, 2019, pp. 141--154.

\bibitem{erekat2020enforcing}
D.~Erekat, Z.~Hammal, M.~Siddiqui, and H.~Dibeklio{\u{g}}lu, ``Enforcing multilabel consistency for automatic spatio-temporal assessment of shoulder pain intensity,'' in \emph{Companion Publication of the 2020 International Conference on Multimodal Interaction}, 2020, pp. 156--164.

\bibitem{xu2020exploring}
X.~Xu and V.~R. de~Sa, ``Exploring multidimensional measurements for pain evaluation using facial action units,'' in \emph{2020 15th IEEE International Conference on Automatic Face and Gesture Recognition (FG 2020)}.\hskip 1em plus 0.5em minus 0.4em\relax IEEE, 2020, pp. 786--792.

\bibitem{aung2015automatic}
M.~S. Aung, S.~Kaltwang, B.~Romera-Paredes, B.~Martinez, A.~Singh, M.~Cella, M.~Valstar, H.~Meng, A.~Kemp, M.~Shafizadeh \emph{et~al.}, ``The automatic detection of chronic pain-related expression: requirements, challenges and the multimodal emopain dataset,'' \emph{IEEE transactions on affective computing}, vol.~7, no.~4, pp. 435--451, 2015.

\bibitem{velana2017senseemotion}
M.~Velana, S.~Gruss, G.~Layher, P.~Thiam, Y.~Zhang, D.~Schork, V.~Kessler, S.~Meudt, H.~Neumann, J.~Kim \emph{et~al.}, ``The senseemotion database: A multimodal database for the development and systematic validation of an automatic pain-and emotion-recognition system,'' in \emph{Multimodal Pattern Recognition of Social Signals in Human-Computer-Interaction: 4th IAPR TC 9 Workshop, MPRSS 2016, Cancun, Mexico, December 4, 2016, Revised Selected Papers 4}.\hskip 1em plus 0.5em minus 0.4em\relax Springer, 2017, pp. 127--139.

\bibitem{gruss2019multi}
S.~Gruss, M.~Geiger, P.~Werner, O.~Wilhelm, H.~C. Traue, A.~Al-Hamadi, and S.~Walter, ``Multi-modal signals for analyzing pain responses to thermal and electrical stimuli,'' \emph{JoVE (Journal of Visualized Experiments)}, no. 146, p. e59057, 2019.

\bibitem{werner2019automatic}
P.~Werner, D.~Lopez-Martinez, S.~Walter, A.~Al-Hamadi, S.~Gruss, and R.~W. Picard, ``Automatic recognition methods supporting pain assessment: A survey,'' \emph{IEEE Transactions on Affective Computing}, vol.~13, no.~1, pp. 530--552, 2019.

\bibitem{cho2014learning}
K.~Cho, B.~Van~Merri{\"e}nboer, C.~Gulcehre, D.~Bahdanau, F.~Bougares, H.~Schwenk, and Y.~Bengio, ``Learning phrase representations using rnn encoder-decoder for statistical machine translation,'' \emph{arXiv preprint arXiv:1406.1078}, 2014.

\bibitem{girard2014nonverbal}
J.~M. Girard, J.~F. Cohn, M.~H. Mahoor, S.~M. Mavadati, Z.~Hammal, and D.~P. Rosenwald, ``Nonverbal social withdrawal in depression: Evidence from manual and automatic analyses,'' \emph{Image and vision computing}, vol.~32, no.~10, pp. 641--647, 2014.

\bibitem{jaiswal2017automatic}
S.~Jaiswal, M.~F. Valstar \emph{et~al.}, ``Automatic detection of adhd and asd from expressive behaviour in rgbd data,'' in \emph{2017 12th IEEE International Conference on Automatic Face \& Gesture Recognition (FG 2017)}.\hskip 1em plus 0.5em minus 0.4em\relax IEEE, 2017, pp. 762--769.

\bibitem{bishay2019schinet}
M.~Bishay, P.~Palasek \emph{et~al.}, ``Schinet: Automatic estimation of symptoms of schizophrenia from facial behaviour analysis,'' \emph{IEEE Transactions on Affective Computing}, 2019.

\bibitem{bishay2019can}
M.~Bishay, S.~Priebe, and I.~Patras, ``Can automatic facial expression analysis be used for treatment outcome estimation in schizophrenia?'' in \emph{ICASSP 2019-2019 IEEE International Conference on Acoustics, Speech and Signal Processing (ICASSP)}.\hskip 1em plus 0.5em minus 0.4em\relax IEEE, 2019, pp. 1632--1636.

\bibitem{wang_17_match}
S.~Wang and J.~Jiang, ``A compare-aggregate model for matching text sequences,'' in \emph{5th International Conference on Learning Representations, {ICLR} 2017, Toulon, France, April 24-26, 2017, Conference Track Proceedings}, 2017.

\bibitem{vinyals_16_oneshot}
O.~Vinyals, C.~Blundell, T.~Lillicrap, K.~Kavukcuoglu, and D.~Wierstra, ``Matching networks for one shot learning,'' in \emph{Proceedings of the 30th International Conference on Neural Information Processing Systems}, ser. NIPS'16, 2016, pp. 3637--3645.

\bibitem{snell_17_prototypical}
J.~Snell, K.~Swersky, and R.~Zemel, ``Prototypical networks for few-shot learning,'' in \emph{Advances in Neural Information Processing Systems}, 2017.

\bibitem{szczapa2022automatic}
B.~Szczapa, M.~Daoudi, S.~Berretti, P.~Pala, A.~Del~Bimbo, and Z.~Hammal, ``Automatic estimation of self-reported pain by trajectory analysis in the manifold of fixed rank positive semi-definite matrices,'' \emph{IEEE transactions on affective computing}, vol.~13, no.~4, pp. 1813--1826, 2022.

\bibitem{shrout1979intraclass}
P.~E. Shrout and J.~L. Fleiss, ``Intraclass correlations: uses in assessing rater reliability.'' \emph{Psychological bulletin}, vol.~86, no.~2, p. 420, 1979.

\bibitem{bishay2020automatic}
M.~A.~T. Bishay, ``Automatic facial expression analysis in diagnosis and treatment of schizophrenia,'' Ph.D. dissertation, Queen Mary University of London, 2020.

\end{thebibliography}
}

\end{document}